\documentclass{article}

\usepackage{microtype}
\usepackage{graphicx}
\usepackage{booktabs}
\usepackage{hyperref}
\usepackage{amsmath}
\usepackage{amssymb}
\usepackage{amsthm}
\usepackage[capitalize]{cleveref}

\usepackage[accepted]{icml2026}

\icmltitlerunning{Towards GFMs for Network Dynamics}

\begin{document}

\twocolumn[
  \icmltitle{
    Towards Graph Foundation Models for Dynamics in Complex Networked Systems:
    Lessons from Super-Spreader Identification in Multilayer Networks
  }

  \icmlsetsymbol{equal}{*}

  \begin{icmlauthorlist}
    \icmlauthor{Micha{\l} Czuba}{equal,wust}
    \icmlauthor{Mateusz Stolarski}{equal,wust}
    \icmlauthor{Adam Pir{\'o}g}{sem}
    \icmlauthor{Piotr Bielak}{wust}
    \icmlauthor{Piotr Br{\'o}dka}{wust}
  \end{icmlauthorlist}

  \icmlaffiliation{wust}{Department of Artificial Intelligence,
    Wroc{\l}aw University of Science and Technology, Wroc{\l}aw, Poland}
  \icmlaffiliation{sem}{4Semantics, Warsaw, Poland}

  \icmlcorrespondingauthor{Micha{\l} Czuba}{michal.czuba@pwr.edu.pl}

  \icmlkeywords{
    graph foundation models,
    inductive learning,
    multilayer networks,
    network dynamics,
    spreading processes
  }

  \vskip 0.3in
]

\printAffiliationsAndNotice{\icmlEqualContribution}

\begin{abstract}
Network dynamics --- including spreading, influence maximisation, and epidemic modelling --- remain largely confined to the transductive paradigm, where models are trained on a single network and cannot be reused on unseen graphs without retraining. We argue that inductive cross-network generalisation is a necessary prerequisite for Graph Foundation Models (GFMs) in this domain and propose four design properties towards this goal. As a proof of concept, \textit{ts-net} (TopSpreadersNetwork), trained solely on synthetic multilayer networks (MLNs), demonstrates zero-shot generalisation to real-world MLNs of varying size and layer count, outperforming classical heuristics and transductive baselines on three of four metrics. Based on \textit{ts-net}'s performance, we further outline five open challenges towards building GFMs for network dynamics: scale, many-layer generalisation, self-supervised pretraining, cross-task transfer, and node-attribute integration.
\end{abstract}

\section{Introduction \label{sec:intro}}

Graph Foundation Models (GFMs) have transformed the graph learning domain, with a growing body of work demonstrating zero-shot and few-shot transfer across node and graph classification tasks. The graph machine learning community is beginning to follow suit, with models such as OFA~\cite{liu2024ofa} and GraphGPT~\cite{tang2024graphgpt} providing early evidence of cross-graph generalisation. One corner of the graph machine learning landscape, however, remains largely rooted in the transductive paradigm: \emph{network dynamics} --- including spreading processes, influence maximisation, and epidemic modelling~\cite{magnani2015spreadingmln}. These tasks are still primarily solved by fitting one model per dataset, per graph, and per task, with only limited progress towards transferable or inductive solutions.

Multilayer networks (MLNs)~\cite{kivela2014multilayer} constitute a powerful framework for modelling complex systems with multiple types of interactions, such as social networks with different communication channels~\cite{jankowski2024timik}, transportation networks with various modes of transit~\cite{cardillo2013eutransportation}, or brain connectivity networks with distinct functional layers~\cite{wilson2021analysis}. Formally, an MLN is a quadruple $G = ([n],[\ell],V,E)$, where $[n]$ is a set of actors, $[\ell]$ a set of layers, $V \subseteq [n] \times [\ell]$ a set of nodes, and $E \subseteq \bigcup_{i \in [\ell]} V_i \times V_i$ a set of edges; each layer $G_i = (V_i, E_i)$ captures a distinct mode of interaction amongst the same actor set. Their rich structure challenges cross-dataset generalisation: layer counts and types vary across datasets, so that even within-domain transfer is non-trivial --- a model trained on a social network with two layers cannot be straightforwardly applied as-is to a transportation network with thirty-seven layers.

In this position paper, we draw on our experience with super-spreader identification in MLNs to argue that \textbf{inductive generalisation across networks, while a necessary prerequisite for any GFM, remains largely unmet in the domain of network dynamics}. We provide both a conceptual argument for why this is so and empirical evidence that it is achievable today.

\section{The transductive status quo \label{sec:status-quo}}

The dominant paradigm for network dynamics tasks is transductive: a model is fitted to a specific graph and cannot be reused on unseen networks without retraining. This limitation is structural rather than incidental, and it cuts across all classes of existing methods.

Classical heuristics like degree centrality (\textit{deg-c}), neighbourhood size (\textit{nghb-s}), and their discounting variants~\cite{chen2009efficient, czuba2024rank}, carry no learnable parameters and accumulate no knowledge across graphs. They apply fixed formulae to each network's adjacency structure independently and cannot improve with more data.

Learning-based methods are transductive for a different but equally fundamental reason: they learn representations tied to a specific graph, whether that representation captures its spreading dynamics, its embedding space, or its centrality structure. Compounding this, most operate on single-layer graphs and cannot directly handle the multilayer setting. \textit{deep-im}~\cite{ling2023deepim} takes the conceptually correct step of internalising spreading dynamics as a learnable, per-graph surrogate (the analogue, at graph scale, of what a GFM spreading model should do at corpus scale) yet re-learns that surrogate from scratch for each new network. PIANO~\cite{piano2023im} recasts influence maximisation as a reinforcement learning problem but is primarily evaluated within structurally similar networks, without explicitly addressing cross-graph transfer. Inductive methods exist in the single-layer setting~\cite{Panagopoulos2024Oct}, but variable layer counts and layer-type heterogeneity make the multilayer case qualitatively harder. Unsupervised multilayer embedding methods such as Multi-node2vec~\cite{wilson2021analysis} address the MLN structure and can be combined with a prediction head, e.g.\ k-means (\textit{mn2v-km}), but remain per-graph, providing no cross-network transfer.

GBIM~\cite{Yuan2024gbim} is a notable exception in addressing a multiplex graph structure, modelling simultaneous multi-information propagation across concurrent layers; nonetheless, it fits a surrogate model per problem instance and transfers nothing across graphs. Models more broadly applicable to ranking in heterogeneous or directed graph settings, such as GNNRank~\cite{he2022gnnrank}, are typically applied in a transductive setting and additionally require adaptation to handle MLNs.

The cost of this paradigm is concrete: limited knowledge transfer across networks, reduced robustness under structural distribution shift, and (for surrogate-based approaches) limited scalability to very large graphs. This is precisely the per-dataset workflow that GFMs have begun to displace in other domains, and we see no fundamental obstacle to the same shift occurring in network dynamics.

\section{Desiderata for GFMs in MLNs' dynamics \label{sec:desiderata}}

We propose four design properties satisfied by \textit{ts-net} and put forward as a candidate minimal set for GFMs in this domain; they are not sufficient on their own, and \Cref{sec:challenges} addresses what remains beyond them.

\paragraph{\textit{D1}: Relationship-agnostic encoding\label{des:encoding}} Layer and edge types should not require a fixed schema known at training time. \textit{ts-net} achieves this via a shared encoder with no layer-type lookup; schema-dependent encoders fix the vocabulary at training time, precluding transfer to networks with unseen layer types.

\paragraph{\textit{D2}: Size-agnostic inference\label{des:size}} The model should generalise to arbitrary actor and layer counts without retraining or architectural modification. Neighbourhood sampling provides actor-side flexibility; adaptive cross-layer aggregation provides layer-side flexibility, enabling zero-shot evaluation on graphs structurally different from those seen during training.

\paragraph{\textit{D3}: Corpus-based training\label{des:corpus}} A GFM must be trained on a \emph{distribution} of graphs rather than a single graph. Even a corpus of synthetic networks constitutes a form of pretraining set, exposing the model to structural diversity (varying density, layer count, and degree distribution) that single-graph training cannot provide.

\paragraph{\textit{D4}: Topology-focused input\label{des:topology}} When the target dynamic is governed primarily by network structure, as in the Independent Cascade Model (ICM)~\cite{goldenberg2001icm}, node features may introduce graph-specific signals that hinder cross-network transfer. Zero-feature input (i.e., zero or degree-derived features rather than node-specific metadata) can therefore serve as a principled design choice in such settings; we provide empirical evidence in \Cref{sec:tsnet}.

\section{A proof of concept --- \textit{ts-net} \label{sec:tsnet}}

We show that \textit{ts-net}~(\textit{TopSpreadersNetwork}~\cite{czuba2025identifyingsuperspreadersmultilayer}) can be treated as a proof of concept for the four desiderata introduced in \Cref{sec:desiderata}: trained entirely on synthetic graphs, it demonstrates zero-shot generalisation to unseen real-world MLNs of widely varying size.

\textit{ts-net} ranks actors by spreading potential under ICM, augmented with a protocol function $\delta$ that generalises it to the multirelational setting. Two boundary cases of $\delta$, representing opposite ends of the diffusion-difficulty spectrum, are considered: $\text{AND}$ requires an actor to receive activation signals in \emph{all} its layers before becoming active; $\text{OR}$ requires activation in \emph{any} single layer. The architecture comprises a shared GAT+GIN encoder~\cite{velivckovic2017graph,xu2018powerful} applied identically to every layer; actor-side neighbourhood sampling and a soft-attention cross-layer aggregation (\textit{WiseAverage}) jointly support \textit{D2} by enabling inference on networks of arbitrary actor and layer count. Per-node outputs are equivariant to within-layer node ordering (via GNN message-passing), and WiseAverage is invariant to layer ordering. An MLP prediction head for the downstream task completes the pipeline with ${\sim}458{,}000$ parameters in total, with all-zero input feature vectors. Training uses the TopSpreaders dataset: more than $200$ synthetic MLNs generated by Erd\H{o}s--R\'enyi and preferential-attachment models~\cite{sf-model}, with ICM simulated under $40$ configurations $\times$ $40$ repetitions per actor, so that each actor serves as a sole diffusion seed across different spreading regimes; no real-world network is seen during training.

\begin{figure}[!t]
\centering
\includegraphics[width=\columnwidth]{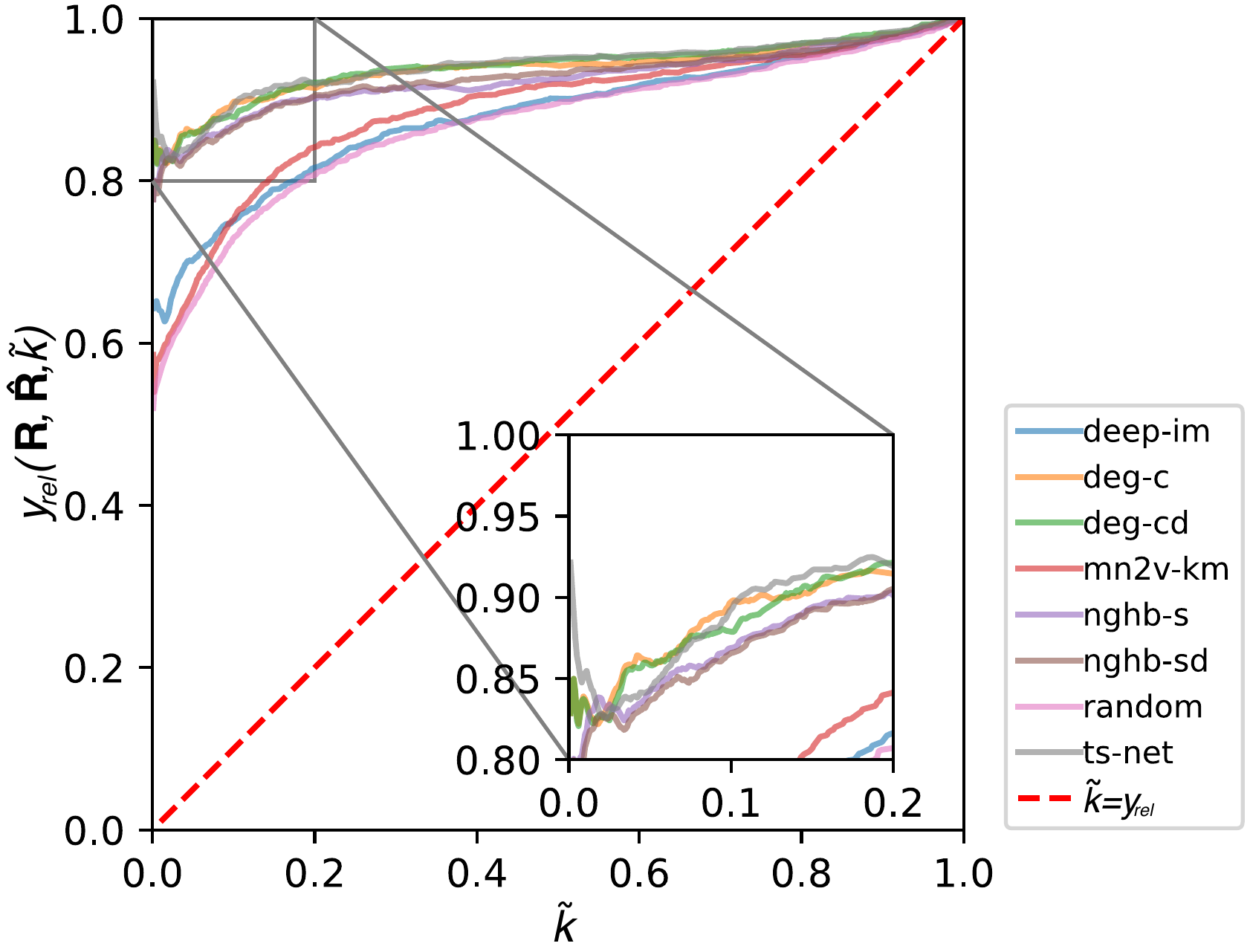}
\caption{Averaged $y_{rel}(k)$ on real-world networks. \textit{ts-net} leads throughout; \textit{deep-im} degrades to near-random performance on most networks.}
\label{fig:curves}
\end{figure}

The spreading potential score of an actor quantifies its capacity to propagate under ICM when seeded alone; $T$ measures that for the top-ranked actor relative to the optimal seed, and $S$ the analogous score for the top-$s$ super-spreaders ($s$ is the 80th-percentile cutoff); both $\in [0,1]$, higher is better~\cite{czuba2025identifyingsuperspreadersmultilayer}. \textit{ts-net} leads in $3$ out of $4$ reported metrics (\Cref{tab:results}), trained only on synthetics yet evaluated zero-shot on real-world MLNs, which provides empirical support for \textit{D3} and \textit{D2}. Notably, \textit{deep-im} could not be evaluated on \textit{timik}~\cite{jankowski2024timik} ($61{,}702$ actors) due to memory exhaustion, a limitation inductive approaches can mitigate. \Cref{fig:curves} traces the cumulative relative score $y_{rel}(k)$ --- the fraction of max. spreading potential recovered by the top-$k$ actors --- across budgets, illustrating \textit{ts-net}'s consistent lead and \textit{deep-im}'s collapse on real-world networks.

\begin{table}[!t]
  \caption{Average $T$ and $S$ across artificial and real-world networks~\cite{czuba2025identifyingsuperspreadersmultilayer}. Bold: best per column.}
  \label{tab:results}
  \vskip 0.1in
  \centering
  \begin{small}
    \begin{tabular}{lcccc}
      \toprule
      & \multicolumn{2}{c}{Artificial} & \multicolumn{2}{c}{Real} \\
      \cmidrule(lr){2-3}\cmidrule(lr){4-5}
      Method & $T$ & $S$ & $T$ & $S$ \\
      \midrule
      random           & 0.538 & 0.759 & 0.520 & 0.749 \\
      \textit{deg-c}   & 0.841 & 0.872 & 0.829 & 0.896 \\
      \textit{deg-cd}  & 0.830 & 0.872 & 0.829 & \textbf{0.901} \\
      \textit{mn2v-km} & 0.611 & 0.790 & 0.586 & 0.786 \\
      \textit{deep-im}  & 0.548 & 0.761 & 0.643 & 0.788 \\
      \textit{ts-net}  & \textbf{0.850} & \textbf{0.881} & \textbf{0.921} & 0.897 \\
      \bottomrule
    \end{tabular}
  \end{small}
\end{table}

\textit{ts-net}, trained exclusively on $\text{AND}$, achieves $T=0.982$ (artificial) and $0.940$ (real) when evaluated under $\text{OR}$ --- \emph{exceeding} the in-distribution $\text{OR}{\to}\text{OR}$ performance of $0.955$ / $0.924$ (\Cref{tab:transfer}). This cross-regime transfer suggests the model captures structural patterns rather than overfitting to regime-specific diffusion behaviour, corroborating \textit{D4}.

\begin{table}[!t]
  \caption{Cross-regime transfer: \textit{ts-net} under different train/test $\delta$ configurations~\cite{czuba2025identifyingsuperspreadersmultilayer}.}
  \label{tab:transfer}
  \vskip 0.1in
  \centering
  \begin{small}
    \begin{tabular}{cccc}
      \toprule
      Train $\delta$ & Test $\delta$ & Art.\ $T$ & Real $T$ \\
      \midrule
      $\text{AND}$ & $\text{AND}$ & 0.808 & 0.899 \\
      $\text{OR}$ & $\text{OR}$  & 0.955 & 0.924 \\
      $\text{AND}$ & $\text{OR}$  & 0.982 & 0.940 \\
      \bottomrule
    \end{tabular}
  \end{small}
  \vspace{-5pt}
\end{table}

Replacing zero-feature input with centrality features reduces $T$ from $0.808\to0.764$ (artificial) and $0.899\to0.730$ (real), providing empirical support for \textit{D4}. In this setting, where dynamics are driven purely by network structure, node features introduce graph-specific noise, and topology alone is sufficient for cross-network transfer.

\section{Open challenges \label{sec:challenges}}

While the results of \textit{ts-net} provide encouraging evidence for the feasibility of inductive generalisation across MLNs in the domain of spreading dynamics, several obstacles need to be addressed before GFMs become practical in this field.

\paragraph{Scale} The training corpus used here (on the order of $10^2$ synthetic networks) is only a first step. Foundation models in other domains rely on orders of magnitude more data. Achieving comparable scale will require both richer synthetic graph generators and the systematic curation of large, diverse repositories of real-world multilayer networks.

\paragraph{Many-layer generalisation} \textit{ts-net}'s performance analysis revealed that performance degrades on networks with substantially more relationships than those seen during training (training graphs had at most $5$ layers; arxiv and eu-trans have $13$ and $37$)~\cite{czuba2025identifyingsuperspreadersmultilayer}. This suggests that actor-side scalability via neighbourhood sampling must be complemented by layer-side strategies, such as chunk-based or hierarchical processing of layers.

\paragraph{Self-supervised pretraining} The discussed approach has been evaluated in a fully supervised manner and depends on costly simulations. Self-supervised objectives, for example contrastive learning or diffusion-prediction surrogates defined on unlabelled graph corpora, could provide a scalable alternative and reduce reliance on expensive labels.

\paragraph{Cross-task transfer} \textit{ts-net} is specialised to super-spreader ranking under ICM. A true GFM should support transfer across related tasks, including source detection, epidemic forecasting, alternative diffusion models (e.g., LTM), and seed set evaluation.

\paragraph{Rich node attributes} Real-world networks often include textual, tabular, or temporal node attributes. Integrating such signals without sacrificing cross-network generalisation remains an open question. While topology-only input may be preferable in settings where the dynamics are purely structural, combining structural and attribute-based information in a transferable manner remains an important direction.

\section{Conclusion \label{sec:conclusion}}

In this article, we argue that inductive generalisation across networks, while a necessary prerequisite for any GFM, remains largely unmet in the domain of network dynamics, and show, through \textit{ts-net}, that such behaviour is achievable in practice, even in the more challenging setting of MLNs. We hope this work motivates the development of scalable, pre-trained, and transferable models for dynamics in complex networked systems, bringing this domain closer to the foundation model paradigm.

\bibliography{references}
\bibliographystyle{icml2026}

\end{document}